%% file: root.tex
\documentclass[letterpaper, 10 pt, conference]{ieeeconf}  

\IEEEoverridecommandlockouts                              

\usepackage{amsmath} 
\usepackage{amssymb} 

\usepackage{multicol}
\usepackage{graphicx}
\usepackage{comment}
\usepackage{bbm}
\usepackage{hyperref}
\usepackage{array}
\usepackage{caption}
\usepackage{algorithm}
\usepackage[noend]{algpseudocode}
\usepackage{xcolor}
\usepackage{flushend}

\usepackage{colortbl}
\definecolor{Gray}{gray}{0.85}
\newcolumntype{g}{>{\columncolor{Gray}}c}

\usepackage{listings}
\newcommand{\pr}[1]{\ensuremath{P\left(#1\right)}}
\newcommand{\pddl}[1]{\texttt{#1}}
\newcommand{\pddlkw}[1]{\textbf{\pddl{#1}}}
\newcommand{\method}[1]{\textsc{#1}}

\title{\LARGE \bf
Online Replanning in Belief Space for Partially Observable \\ 
Task and Motion Problems
}

\author{Caelan Reed Garrett$^{1,2}$, Chris Paxton$^{2}$, Tom\'as Lozano-P\'erez$^{1}$, Leslie Pack Kaelbling$^{1}$, Dieter Fox$^{2}$
\thanks{$^{1}$MIT CSAIL,{\tt\small \{caelan,tlp,lpk\}@csail.mit.edu}}%
\thanks{$^{2}$NVIDIA,{\tt\small \{cpaxton,dieterf\}@nvidia.com}}%
\thanks{We gratefully acknowledge support from NSF grants 1523767 and 1723381; from AFOSR grant FA9550-17-1-0165; from ONR grant N00014-18-1-2847; from the Honda Research Institute; and from SUTD Temasek Laboratories.  Any opinions, findings, and conclusions or recommendations expressed in this material are those of the authors and do not necessarily reflect the views of our sponsors. 
}
}

\usepackage{color}
\usepackage{xcolor}

\begin{document}


\maketitle
\thispagestyle{empty}
\pagestyle{empty}



\begin{abstract}
To solve multi-step manipulation tasks in the real world, an autonomous robot must take actions to observe its environment and react to unexpected observations.
This may require opening a drawer to observe its contents or 
moving an object out of the way to examine the space behind it.
Upon receiving a new observation, the robot must update its belief about the world and compute a new plan of action.  
In this work, we present an online planning and execution system for robots faced with these challenges. 
We perform deterministic cost-sensitive planning in the space of hybrid belief states to select likely-to-succeed observation actions and continuous control actions.
After execution and observation, we replan using our new state estimate. 
We initially enforce that planner reuses the structure of the unexecuted tail of the last plan.
This both improves planning efficiency and ensures that the overall policy does not undo its progress towards achieving the goal.
Our approach is able to efficiently solve partially observable problems both in simulation and in a real-world kitchen.

\end{abstract}


\section{Introduction}

Robots acting autonomously in human environments are faced with a variety of challenges.
First, they must make both discrete decisions about what object to manipulate as well as continuous decisions about which motions to execute to achieve a desired interaction.
Planning in these large {\em hybrid} spaces is the subject of integrated {\em Task and Motion Planning} (TAMP)~\cite{gravot2005asymov,HPN,srivastava2014combined,toussaint2015logic,garrettIJRR2017,garrettIJRR2018}.
Second, real-world robot actions are often quite {\em stochastic}.
Uncertainty in the effects of actions can manifest both locally due noisy continuous actuation or more broadly due to unexpected changes in contact.
Third, the robot can only {\em partially observe} the world due to occlusions caused by doors, drawers, other objects, and even the robot itself.
Thus, the robot must maintain a {\em belief} over the locations of entities and intentionally select actions that reduce its uncertainty about the world~\cite{IJRRBel}.

This class of problems can be formalized as a hybrid {\em partially observable Markov decision process} (POMDP)~\cite{kaelbling1998planning}.
Solutions are {\em policies}, mappings from distributions over world states ({\em belief-states}) to actions.
Because solving these problems exactly is intractable~\cite{kaelbling1998planning}, we compute a policy {\em online} via repeatedly {\em replanning}~\cite{FFReplan,kolobov2012planning},
each time solving an approximate, {\em determinized}~\cite{FFReplan,kolobov2012planning} version of the problem using an existing TAMP approach~\cite{garrett2020PDDLStream}.
POMDP planning can be viewed as searching through {\em belief space}, the space of belief states, where perception and control actions operate on distributions over states instead of individual states.

Most prior work approximately models belief space using either discrete~\cite{phiquepal2017combined,wang2018bounded,srivastava2018anytime} or fluent-based~\cite{IJRRBel,hadfield2015modular} {\em abstractions}.
In contrast, we operate {\em directly} on hybrid belief distributions by specifying procedures that model observation sampling, visibility checking, and Bayesian belief filtering.
This allows us to tackle problems where continuous and geometric components of the state govern the probability of an observation.
For example, a movable object at a particular pose might occlude a goal object, substantially reducing the probability that it will be detected.
By using a {\em particle-based} belief representation, we can model multi-modal beliefs that arise when several objects occlude regions of space.
During planning, we conservatively approximate the probability of detection by {\em factoring} it into a conjunction over conditions on each individual object.
This exposes {\em sparse} interactions between an observation and each object, enabling the planner to efficiently reason about and rectify occlusion. 

\begin{figure}[bt]
  \centering
    \includegraphics[width=\columnwidth]{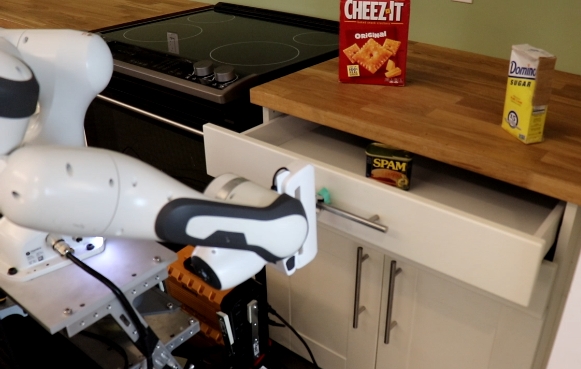}
    \caption{The robot pulls open a drawer to detect whether the spam object is at a continuous pose particle within the drawer.
    }
    \label{fig:kitchen}
    \vskip -0.3cm
\end{figure}

Additionally, we introduce a replanning algorithm that uses past plans to {\em constrain} the structure of solutions for the current planning problem.
This forces future plans to retain the structure of prior plans but allows some of the parameter values to change in response to stochastic execution or new observations.
As a result, this ensures that the overall induced policy continues to make {\em progress} towards achieving the goal.
Additionally, this reduces the search space of the planner and thus speeds up successive replanning invocations.

Finally, we introduce a mechanism that {\em defers} computing values for plan parameters that 1) are used temporally later in the plan, 2) almost always admit satisfying values, and 3) are expensive to compute.
Intuitively, this strategy performs the least amount of computation possible to obtain the next action and certify that it will make progress towards the goal. 
Finally, we evaluate our algorithms on several simulated tasks and demonstrate our system running on a real robot acting in a kitchen environment in the accompanying video.
\section{Related Work}




There is much work that addresses the problem of efficiently solving {\em deterministic}, {\em fully-observable} TAMP problems~\cite{gravot2005asymov,HPN,srivastava2014combined,toussaint2015logic,garrettIJRR2017,garrettIJRR2018}.
However, only a few of these approaches have been extended to incorporate some level of stochasticity or partial observability~\cite{IJRRBel,hadfield2015modular,phiquepal2017combined}.

Solving for an optimal, closed loop policy is undecidable in the infinite-horizon case~\cite{kaelbling1998planning,madani1999undecidability}, even for discrete POMDPs.
An alternative strategy is to dynamically compute a policy {\em online} in response to the current belief, rather than {\em offline} for all beliefs, by {\em replanning}~\cite{kolobov2012planning}.
One approach to online planning is to use Monte-Carlo sampling~\cite{silver2010monte,somani2013despot} to explore likely outcomes of various actions.
These methods have been successfully applied to robotic planning tasks such as grasping in clutter~\cite{li2016act}, non-prehensile rearrangement~ \cite{king2017unobservable}, and object search~\cite{xiao2019online}.
However, the hybrid action space in our application is too high-dimensional for uninformed action sampling to generate useful actions.

Another online planning strategy is to approximate the original stochastic problem as a deterministic problem through the process of {\em determinization}~\cite{FFReplan,yoon2008probabilistic,kolobov2012planning}. 
This enables deterministic planners, which are able to efficiently search large spaces, to be applied.
{\em Most-likely outcome} determinization always assigns the action outcome that has the highest probability.
When applied to observation actions, this approach is called {\em maximum likelihood observation} (MLO) determinization~\cite{Platt2010BeliefObservations,platt2012non,hadfield2015modular}.
However, this approximation fails when the success of a policy depends on some outcome other than the mode of observation distribution.

There are many approaches for representing and updating a belief
such as joint, unscented Kalman filtering~\cite{Platt2010BeliefObservations,IJRRBel},
{\em factoring} the belief into independent distributions per object~\cite{hadfield2015modular,kaelbling2016implicit}, and maintaining a particle filter,  which represents the belief as a set of weighted samples~\cite{silver2010monte,somani2013despot,li2016act,xiao2019online}.
Many approaches use a different belief representation when planning versus when filtering.
Several approaches plan on a purely discrete {\em abstraction} of the underlying hybrid problem~\cite{phiquepal2017combined,wang2018bounded,srivastava2018anytime}.
Other approaches plan using a calculus defined on {\em belief fluents}~\cite{IJRRBel,hadfield2015modular}, logical tests on the underlying belief such as ``the value of random variable $X$ is within $\delta$ of value $x$ with probability at least $1 - \epsilon$''.
In contrast, our approach plans directly on probability distributions, where actions update beliefs via proper transition and observation updates.
\section{Problem Definition}

We address hybrid, belief-state {\em Stochastic Shortest Path Problems} (SSPP)~\cite{bertsekas1991analysis}, a subclass of hybrid POMDPs where 
the {\em cost} $c_a > 0$ of action $a$ is strictly positive.
The robot starts with a {\em prior} belief $b_0$. 
Its objective is to reach a goal set of beliefs $B_*$ while 
minimizing the cost it incurs.
The robot selects actions $a$ according to a {\em policy} $a \sim \pi(b)$ defined on belief states $b$.
We evaluate $\pi(b)$ {\em online} by {\em replanning} given the current belief state $b$.
We approximate the original belief-space SSPP by {\em determinizing} its action outcomes (Section~\ref{sec:cost}).
We formalize each determinized SSPP in the PDDLStream~\cite{garrett2020PDDLStream} language and solve them using a cost-minimizing PDDLStream planner.



\begin{figure}[bt]
  \centering
    \includegraphics[width=0.27\textwidth]{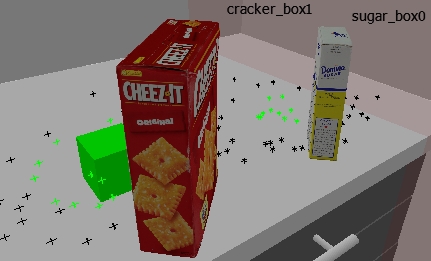}
    \includegraphics[width=0.205\textwidth]{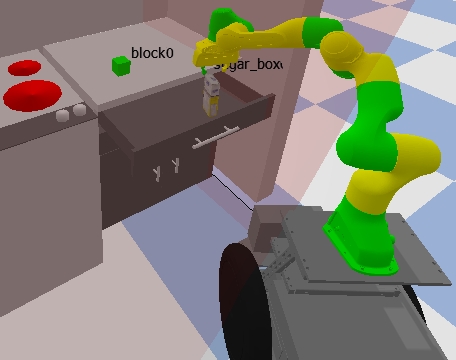}
    \caption{\textbf{Left}: the particle-filter pose belief for the green block after one observation. Green particles have high weight and black particles have low weight. \textbf{Right}: in the {\em Stow} task, the robot must remove the sugar to place the block and close the drawer.}
    \label{fig:particle}
    \vskip -0.3cm
\end{figure}

Although our technique is general-purpose, our primary application is partially-observable TAMP in a kitchen environment that contains a single mobile manipulator, counters, cabinets, drawers, and a set of unique, known objects.
The robot can observe the world using an RGB-D camera that is fixed to the world frame.
The camera can {\em detect} the set of objects that are visible as well as noisily {\em estimate} their poses.
The latent world state is given by the robot configuration, door and drawer joint angles, the discrete frame that each object is attached to, and the pose of each object relative to its attached frame.
We maintain a {\em factored} belief as the product of independent posterior distributions over each variable.
In our environment, the robot's configuration as well as the door and drawer joint angles can be accurately estimated using our perception system~\cite{schmidt2014dart}, so we only maintain a point estimate for these variables.
However, there is substantial partial observability when estimating object poses due to occlusions from doors, drawers, other objects, and even the robot.
We represent and update our belief over the pose state of each object using {\em particle filtering} (Fig.~\ref{fig:particle}).


\section{PDDLStream Formulation}




PDDLStream is an extension of Planning Domain Definition Language (PDDL)~ \cite{mcdermott1998pddl} 
that adds the ability to programmatically declare procedures for sampling values of continuous variables in the form of {\em streams}.
Like PDDL, PDDLStream uses {\em predicate} logic to describe planning problems. 
An evaluation of a predicate for a given set of arguments is called a {\em literal}. 
A {\em fact} is a true literal.
{\em Static} literals always remain constant, but {\em fluent} literals can change truth value as actions are applied.
States are represented as a set of fluent literals.
Our domain makes use of the following fluent predicates: 
\pddl{(AtConf ?r ?q)} states that robot part \pddl{?r} (the \pddl{base} or \pddl{arm}) is at configuration \pddl{?q};
\pddl{(AtAngle ?j ?a)} states that a door or drawer \pddl{?j} is at joint angle \pddl{?a}; 
\pddl{(HandEmpty)} indicates that the robot's end-effector is empty; and
\pddl{(AtGrasp ?o ?g)} states that object \pddl{?o} is attached to the end-effector using grasp \pddl{?g}.
Here, \pddl{?q}, \pddl{?a}, and \pddl{?g} are all real-valued high-dimensional parameters.


An {\em action schema} is specified by a set of free parameters (\pddlkw{:param}), a precondition formula (\pddlkw{:pre}) that must hold  in a state in order to execute the action, and a conjunctive effect formula (\pddlkw{:eff}) that describes the changes to the state.
Effect formulas may set a fluent fact to be true, set a fluent fact to be false (\pddlkw{not}), or increase the plan cost (\pddlkw{incr})~\cite{Fox03pddl2.1:an}.
For example, consider the following action descriptions for \pddl{move} and \pddl{pick}.
Other actions such as \pddl{place}, \pddl{push} door, \pddl{pull} door, \pddl{pour}, and \pddl{press} button can be defined similarly to \pddl{pick}. 
We used universally quantified {\em conditional effects}~\cite{pednault1989adl} (omitted here for clarity) to update the world poses of objects placed in drawers for \pddl{push} and \pddl{pull} actions.

\begin{footnotesize}
\begin{lstlisting}
(|\textbf{:action}| move
 |\textbf{:param}| (?r ?q1 ?t ?q2)
 |\textbf{:pre}| (|\textbf{and}| (Motion ?r ?q1 ?t ?q2) (AtConf ?r ?q1))
 |\textbf{:eff}| (|\textbf{and}| (AtConf ?r ?q2) (|\textbf{not}| (AtConf ?r ?q1))))
(|\textbf{:action}| pick
 |\textbf{:param}| (?o ?pb ?g ?bq ?aq)
 |\textbf{:pre}| (|\textbf{and}| (Kin ?o ?pb ?g ?bq ?aq) (AtPoseB ?o ?pb) 
  (HandEmpty) (AtConf base ?bq) (AtConf arm ?aq))
 |\textbf{:eff}| (|\textbf{and}| (Holding ?o ?g)
  (|\textbf{not}| (AtPoseB ?o ?pb)) (|\textbf{not}| (HandEmpty))))
\end{lstlisting}
\end{footnotesize}





The novel representational aspect of PDDLStream is {\em streams}: functions from a set of {\em input} values (\pddlkw{:inp}) to a {\em generator} that enumerates a possibly infinitely-long sequence of {\em output} values (\pddlkw{:out}).
Streams have a {\em declarative} component that specifies 1) arity of input and output values, 2) a {\em domain formula} (\pddlkw{:dom}) that governs legal inputs, and 3) a conjunctive {\em certified formula} (\pddlkw{:cert}) that expresses static facts that all input-output pairs are {\em guaranteed} to satisfy.
Additionally, streams have a {\em programmatic} component that implements the function in a programming language such as Python.
For example, the \pddl{inv-kin} stream takes in a tuple of values specifying an object \pddl{?o}, its pose \pddl{?pb}, a grasp \pddl{?g}, and a robot base configuration \pddl{?bq}.
Using an inverse kinematics solver, it generates robot arm configurations \pddl{?aq} that satisfy the attachment kinematic relationship \pddl{Kin} involving the provided placement and grasp for the object.
The \pddl{motion} stream uses a motion planner, such as RRT-Connect~\cite{KuffnerLaValle}, to produce trajectories \pddl{?t} that certify the static  \pddl{Motion} precondition of the \pddl{move} action.

\vspace{-1ex}
\begin{footnotesize}
\begin{multicols}{2}
\begin{lstlisting}
(|\textbf{:stream}| inv-kin
 |\textbf{:inp}| (?o ?pb ?g ?bq)
 |\textbf{:dom}| (|\textbf{and}| (Conf base ?bq) 
  (PoseB ?o ?pb) (Grasp ?o ?g)
 |\textbf{:out}| (?aq)
 |\textbf{:cert}| (|\textbf{and}| (Conf arm ?aq) 
  (Kin ?o ?p ?g ?bq ?aq)))
\end{lstlisting}
\begin{lstlisting}
(|\textbf{:stream}| motion
 |\textbf{:inp}| (?r ?q1 ?q2)
 |\textbf{:dom}| (|\textbf{and}| (Conf ?r ?q1)
           (Conf ?r ?q2))
 |\textbf{:out}| (?t)
 |\textbf{:cert}| (|\textbf{and}| (Traj ?r ?t) 
  (Motion ?r ?q1 ?t ?q2)))
\end{lstlisting}
\end{multicols}
\end{footnotesize}
\vspace{-2em}



\subsection{Modeling Observations} \label{sec:observations}

Our first contribution is applying PDDLStream~\cite{garrett2020PDDLStream} to model determinized, hybrid belief-state SSPPs.
This allows us to plan using {\em domain-independent} PDDLStream algorithms, such as the {\em Focused} algorithm~\cite{garrett2020PDDLStream}, without modification.
We model the ability for the robot to perform a sensing action, receive an observation, and update its belief using the \pddl{detect} action.
The \pddl{detect} action is parameterized by an object \pddl{?o}, a {\em prior pose belief} \pddl{?pb1}, an {\em observation} \pddl{?obs}, and a {\em posterior pose belief} \pddl{?pb2}. 
The fluent \pddl{(AtPoseB ?o ?pb)} states that object \pddl{?o} has current pose belief \pddl{?pb}.
Critically, \pddl{?pb1} and \pddl{?pb2} represent {\em distributions} over real-valued poses.
The \pddl{BeliefUpdate} precondition ensures these values represent a Bayesian update.
If the observation \pddl{?obs} is not \pddl{BOccluded} by another object, \pddl{detect} updates the current pose belief for \pddl{?o}.

\begin{footnotesize}
\begin{lstlisting}
(|\textbf{:action}| detect
 |\textbf{:param}| (?o ?pb1 ?obs ?pb2)
 |\textbf{:pre}| (|\textbf{and}| (BeliefUpdate ?o ?pb1 ?obs ?pb2)
  (AtPoseB ?o ?pb1) (|\textbf{not}| (BOccluded ?o ?pb1 ?obs)))
 |\textbf{:eff}| (|\textbf{and}| (AtPoseB ?o ?pb2) (|\textbf{not}| (AtPoseB ?o ?pb1))
  (|\textbf{incr}| (|\textbf{total-cost}|) (ObsCost ?o ?pb1 ?obs))))
\end{lstlisting}
\end{footnotesize}


Our key representational insight is that we can encode the Bayesian filtering process by defining sampling and inference {\em streams} that operate on distributions. 
The \pddl{sample-obs} stream samples \pddl{?obs} from the distribution of observations using the {\em observation model} for \pddl{?o} and a pose belief \pddl{?pb}.
The \pddl{test-vis} stream returns true if object \pddl{?o2} at belief \pddl{?pb2} prevents the robot from observing \pddl{?obs} with probability less than $\epsilon$, a value described in Section~\ref{sec:cost}.
The probability of occlusion is estimated by performing ray-casting along \pddl{?obs} using poses sampled from \pddl{?pb2}.


\begin{footnotesize}
\begin{lstlisting}
(|\textbf{:stream}| sample-obs (|\textbf{:stream}| test-vis
 |\textbf{:inp}| (?o ?pb)       |\textbf{:inp}| (?o ?obs ?o2 ?pb2)
 |\textbf{:dom}| (PoseB ?o ?pb) |\textbf{:dom}| (|\textbf{and}| (Obs ?o ?obs)
 |\textbf{:out}| (?obs)          (PoseB ?o2 ?pb2))
 |\textbf{:cert}| (Obs ?o ?obs))|\textbf{:cert}| (BVis ?o ?obs ?o2 ?pb2))
\end{lstlisting}
\end{footnotesize}

\noindent
The \pddl{update-belief} stream computes the posterior pose belief \pddl{?pb2} 
by performing a {\em Bayesian update} using the prior pose belief \pddl{?pb1} and hypothesized observation \pddl{?obs}.
Note that although observations are stochastic, 
the belief update process given an observation is a deterministic function. 

\begin{footnotesize}
\begin{lstlisting}
(|\textbf{:stream}| update-belief
 |\textbf{:inp}| (?o ?pb1 ?obs)
 |\textbf{:dom}| (|\textbf{and}| (PoseB ?o ?pb1) (Obs ?o ?obs))
 |\textbf{:out}| (?pb2)
 |\textbf{:cert}| (|\textbf{and}| (PoseB ?o ?pb2)
  (BeliefUpdate ?o ?pb1 ?obs ?pb2)))
\end{lstlisting}
\end{footnotesize}

Finally, we specify \pddl{BOccluded} as a derived predicate~\cite{edelkamp2004pddl2,thiebaux2005defense}, a fact that is logically inferred given the current state.
\pddl{BOccluded} is true if there exists another object \pddl{?o2} currently at pose belief \pddl{?pb2} that prevents \pddl{?obs} from being received with probability exceeding $\epsilon$.

\begin{footnotesize}
\begin{lstlisting}
(|\textbf{:derived}| (BOccluded ?o ?obs)
 (|\textbf{exists}| (?o2 ?pb2)
  (|\textbf{and}| (Obs ?o ?obs)  (AtPoseB ?o2 ?pb2) 
   (|\textbf{not}| (BVis ?o ?obs ?o2 ?pb2)))))  
\end{lstlisting}
\end{footnotesize}
\vspace{-1em} 
%

\section{Determinized Observation Costs}
\label{sec:cost}


We are interested in enabling a deterministic planner to perform approximate probabilistic reasoning by minimizing plan costs.
The maximum acceptable risk can always be specified using a user-provided maximum expected cost $c_* \in [0, \infty)$.
We focus on computing \pddl{ObsCost}, the cost of \pddl{detect}, which is a function of the prior pose belief \pddl{?pb1} and the observation \pddl{?obs}.
Similar analysis can be applied to other probabilistic conditions, such as collision checks.

\begin{footnotesize}
\begin{lstlisting}
(|\textbf{:function}| (ObsCost ?o ?pb ?obs)
 |\textbf{:dom}| (|\textbf{and}| (PoseB ?o ?pb) (Obs ?o ?obs)))
\end{lstlisting}
\end{footnotesize}

\subsection{Self-Loop Determinization}

The widely-used most-likely-outcome and all-outcome determinization schemes do not provide a natural way of integrating the cost $c_a$ of action $a$ and the probability of an intended outcome $p_a$~\cite{blum1999probabilistic, IJRRBel}.
Thus, we instead use {\em self-loop} determinization~\cite{keyder2008hmdpp,kolobov2012planning}, which approximates the original SSPP as a simplified self-loop SSPP.
In a self-loop SSPP, an action $a$ executed from state $s$ may result in only two possible states: a new state $s'$ or the current state $s$.
For this simple class of SSPPs, a planner can obtain an optimal policy by optimally solving a deterministic problem with {\em transformed} action costs.
Let $c_a'$ be the cost of $a$ upon a failed (self-loop) transition.
The determinized cost $\hat{c}_a$ of action $a$ is then
\begin{equation*}
    \hat{c}_a \equiv c_a + \sum_{t=1}^\infty c_a' (1-p_a)^t = c_a + \Big(\frac{c_a'}{p_a} - c_a'\Big).
\end{equation*}
We directly model our domain as a self-loop SSPP by specifying an upper bound for expected cost of a successful outcome $c_a$, an upper bound for the expected {\em recovery} cost $c_a'$ to return to $s$ ({\it i.e.} the self-loop transition), and a lower bound for the probability of a successful outcome $p_a$.


%
%
\subsection{Computing the Likelihood of an Observation.} 

Suppose there are $n$ unique objects in the world, and we are interested in detecting object $i$.
Let ${\cal X}_j$ be the latent continuous pose random variable for an object $j$, and let $x_j$ be a value of ${\cal X}_j$. 
As shorthand, define $\bar{\cal X}_{-i}$ 
to be a tuple of latent poses for each of the $n$ objects {\em except} for object $i$.
Let $\pr{{\cal X}_i}$ be a probability density over ${\cal X}_i$, which in our application, is represented by a set of weighted particles.
Let ${\cal Z}^v_i$ and ${\cal Z}^d_i$ be observed Bernoulli random variables for whether object $i$ is {\em visible} and is {\em detected}.
When ${\cal Z}^d_i$ is true, let ${\cal Z}^p_i$ be a continuous random variable for the observed pose of object $i$. 
Otherwise, ${\cal Z}^p_i$ is undefined. 
For detection, we will assume that $\pr{{\cal Z}^d_i {=} 1 \mid {\cal Z}^v_i {=} 1}  = 1 - p_{\text{FN}}$ where $p_{\text{FN}}$ is the probability of a false negative.
We will conservatively use zero as a lower bound for the probability of a false positive, {\it i.e.} $\pr{{\cal Z}^d_i {=} 1 \mid {\cal Z}^v_i {=} 0} \geq 0$, which removes false detection terms.
For pose observations, we will assume a multivariate Gaussian noise model ${\cal Z}^p_i \mid \big({\cal Z}^d_i {=} 1, {\cal X}_i {=} x_i\big) \sim \mathcal{N}(x_i,\,\Sigma_i)$. 
We are interested in $\pr{{\cal Z}^p_i}$, the probability of receiving a pose observation for object $i$.

\vspace{-1ex}
\begin{small}
\begin{equation*}
    \pr{{\cal Z}^p_i} = \int_{x_i}\pr{{\cal Z}^p_i \mid {\cal Z}^d_i, x_i} \pr{{\cal Z}^d_i \mid {\cal Z}^v_i}\pr{{\cal Z}^v_i \mid x_i}d\pr{x_i}
\end{equation*}
\end{small}
\vspace{-1ex}

The key component of this expression is $\pr{{\cal Z}^v_i \mid x_i}$, the probability that $x_i$ is currently visible, which is contingent on the poses of the other objects $\bar{\cal X}_{-i}$. 
Define ${B^i_j}(x_i, x_j)$ as a deterministic function that is $1$ if object $j$ at pose $x_j$ {\em blocks} object $i$ from being visible at pose $x_i$ and otherwise is $0$.
Ultimately, $\pr{{\cal Z}^v_i \mid x_i}$ will be a component of the cost function \pddl{ObsCost} and thus must only depend on pose belief \pddl{?pb1} and observation \pddl{?obs}.
However, it is currently still dependent on the current beliefs for each of the other $n-1$ objects all at once.
While we could instead parameterize \pddl{ObsCost} using the pose belief of all objects, it would be combinatorially difficult to instantiate as $n$ increases.
And due to its {\em unfactored} form, we would not be able to benefit from efficient deterministic search strategies that leverage factoring.
Thus, we marginalize out $x_i$, which ties the ${n-1}$ objects together, by taking the worst-case probability of visibility $L_j(X_i)$ due to object $j$ over a 
subset of states $X_i$.
\begin{equation*}
    L_j(X_i) = \inf_{x_i \in X_i} \int_{x_j} \big(1-{B^i_j}(x_i, x_j)\big) d\pr{x_j} \label{eqn:bound}
\end{equation*}
As a result, we can provide a non-trivial lower bound for $\pr{{\cal Z}^v_i \mid x_i}$ that no longer depends on $x_i$.
Suppose there exists $\epsilon \in [0, 1)$ that satisfies $\min_{j \neq i} L_j(X_i) \geq \epsilon$, then
\begin{align}
    \pr{{\cal Z}^v_i \mid x_i} &= \int_{\bar{x}_{-i}} (1 - {B^i_{-i}}(x_{i}, \bar{x}_{-i})) d\pr{\bar{x}_{-i}} \notag \\
    &\geq \prod_{j \neq i}^n \int_{x_j} \big(1 - {B^i_j}(x_i, x_j)\big) d\pr{x_j} \label{eqn:factor} \\
    &\geq \big(\min_{j \neq i} L_j(X_i) \big)^{n-1} \geq (1 - \epsilon)^{n-1}. \notag 
\end{align}
Inequality~\ref{eqn:factor} follows from the fact that some combinations of $\bar{x}_{-i}$ would result in object collision and thus are not possible.
Finally, this gives us the following lower bound for $p_a$:
\begin{align}
    \pr{{\cal Z}^p_i} &\geq \pr{{\cal Z}^p_i, {\cal X}_i \in X_i} \notag  \\
    &\geq p_{\text{FN}} \pr{{\cal Z}^p_i, {\cal X}_i \in X_i  \mid {\cal Z}^d_i}  (1-\epsilon)^{n-1} \label{eqn:pobs}
\end{align}

This probability depends on both $X_i$ and $\epsilon$.
Ideally, we would select $X_i$ and $\epsilon$ that maximize equation~\ref{eqn:pobs}; however, this requires operating on all $n$ objects at once.
Instead, we let the planner select $\pddl{?obs} = (z_i^p, X_i, \epsilon)$ by sampling different values of \pddl{?obs}. 
However, \pddl{detect} can only be applied at this cost if $\forall j \neq i,\; L_j(X_i) \geq 1 - \epsilon$, which is enforced using the \pddl{BOccluded} derived predicate (section~\ref{sec:observations}) by quantifying over the \pddl{BVis} condition for each object.
The choice of $X_i$ presents a trade off because the prior probability $\pr{X_i}$ increases as $X_i$ grows but each $L_j(X_i)$ decreases.
In practice, we sample $x_i \sim {\cal X}_i$ and take 
$X_i = \{x_i' \in B_\delta(x_i) \mid 0 < \pr{x_i'}\}$
to be a $\delta$-neighborhood of $x_i$, 
capturing a local region where we anticipate observing object $i$.
\subsection{Observation Example}

\begin{figure}[bt]
  \centering
    \includegraphics[width=0.33\textwidth]{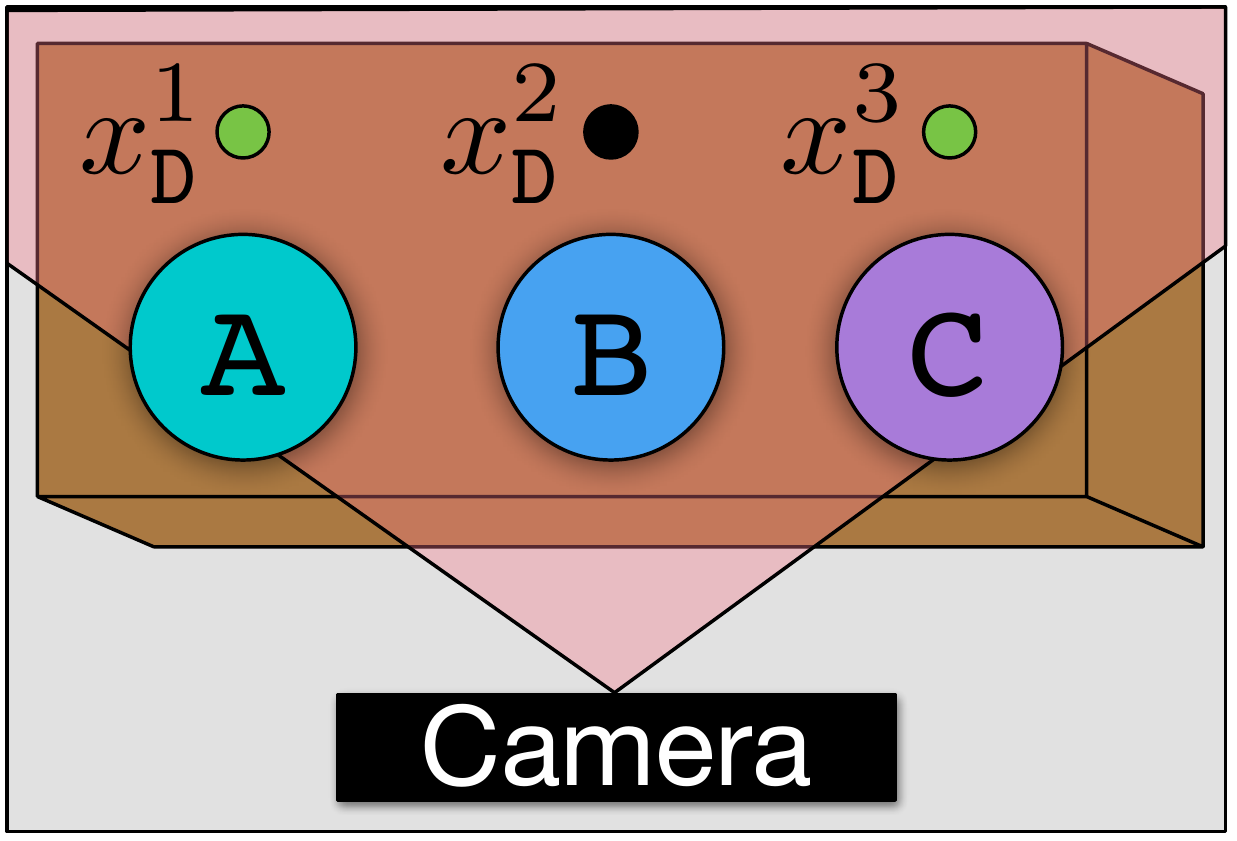}
    \caption{An example detection scenario where object \pddl{D} is believed to be either behind object \pddl{A} or object \pddl{C} with equal probability.} \label{fig:belief}
    \vskip -0.3cm
\end{figure}

Consider the scenario in
Fig.~\ref{fig:belief} with objects \pddl{A}, \pddl{B}, \pddl{C}, and \pddl{D}.
Suppose that the object poses for \pddl{A}, \pddl{B}, and \pddl{C} are perfectly known, but object \pddl{D} is equally believed to be either at pose $x_\pddl{D}^1$ or $x_\pddl{D}^3$ (but not $x_\pddl{D}^2$).
First, note that $L_j(X_i) = 0$ for all choices of $X_i$
because object $\pddl{A}$ occludes $x_\pddl{D}^1$, object $\pddl{B}$ occludes $x_\pddl{D}^2$, and object $\pddl{C}$ occludes $x_\pddl{D}^3$, all with probability 1.
If we take $X_\pddl{D} = \{x_\pddl{D}^1, x_\pddl{D}^2, x_\pddl{D}^3\}$, then $L_\pddl{A}(X_\pddl{D}) = L_\pddl{B}(X_\pddl{D}) = L_\pddl{C}(X_\pddl{D}) = 1$, meaning all three objects must be moved before applying \pddl{detect}, despite the fact that $\pr{x_\pddl{D}^2} = 0$. 
If we take $X_\pddl{D} = \{x_\pddl{D}^1, x_\pddl{D}^3\}$ then $L_\pddl{A}(X_\pddl{A}) = L_\pddl{C}(X_\pddl{D}) = 1$ but $L_\pddl{B}(X_\pddl{A}) = 0$, indicating that $\pddl{B}$ does not need to be moved.
Finally, if we take $X_\pddl{D} = \{x_\pddl{D}'\}$ then only $L_\pddl{A}(X_\pddl{A}) = 1$, and only \pddl{A} must be moved. 
Intuitively, this shows that selecting $X_i$ to be a small, local region improves sparsity with respect to which objects affect a particular observation under our bound.

\section{ONLINE REPLANNING} \label{sec:replanning}

Now that we have incorporated probabilistic reasoning into a deterministic planner, we induce a policy $\pi$ by replanning after executing each action $a$.
However, done naively, it is possible to result in a policy that never reaches the goal set of beliefs $B_*$.
This is even true when acting in a {\em deterministic problem} using replanning.
For example, consider a deterministic, observable planning problem where the goal is for the robot to hold object \pddl{D}.
The robot's first plan might require moving its \pddl{base} (\pddl{b}), moving its \pddl{arm} (\pddl{a}), and finally picking object \pddl{D}: {\small $[\pddl{move}(\pddl{base}, q^0_\pddl{b}, t^1_\pddl{b}, q^1_\pddl{b}),$ 
$\pddl{move}(\pddl{arm}, q^0_\pddl{a}, t^1_\pddl{a}, q^1_\pddl{a}),$
$\pddl{pick}(\pddl{D}, p^0_\pddl{D}, g_\pddl{D}, q^1_\pddl{b}, q^1_\pddl{a})]$}. 
Suppose the robot executes the first \pddl{move} action, arrives as expected at \pddl{base} configuration $q^1_\pddl{b}$, but replans to obtain a new plan: {\small $[\pddl{move}(\pddl{base}, q^1_\pddl{b}, t^2_\pddl{b}, q^2_\pddl{b}),$ $\pddl{move}(\pddl{arm}, q^0_\pddl{a}, t^2_\pddl{a}, q^2_\pddl{a}),$ $\pddl{pick}(\pddl{D}, p^0_\pddl{D}, g_\pddl{D}, q^2_\pddl{b}, q^2_\pddl{a})]$}.


While this is a satisfactory, despite suboptimal, solution when solving for a single plan in isolation, it is undesirable in the context of the previous plan because it requires another \pddl{base} movement action despite the robot having just executed one.
This process could repeat indefinitely, causing the robot to {\em never} achieve its goal despite never failing to find a plan.
For a deterministic problem, this is easily preventable by simply executing the first plan all at once.
However, in a stochastic environment where, for example, base movements are imprecise, executing the full plan open loop will almost always fail.
Thus, we must replan using the base pose $\hat{q}^1_\pddl{b}$ that we actually reach instead of $q^1_\pddl{b}$, the one we intended to reach.
This at least requires planning new values for anything that was a function of $q^1_\pddl{b}$, such as $q^1_\pddl{a}$, which was sampled by the $\pddl{inv-kin}(\pddl{D}, p^0_\pddl{D}, g_\pddl{D}, q^1_\pddl{b})$ stream. 

Intuitively, we need to enforce that some amount of overall {\em progress} is retained when replanning. 
One way to do this is to impose a decreasing {\em constraint} on the length of future plans.
This constraint could be that the next plan must have fewer actions than the previous plan.
If actions always have positive probability of successful execution, for example if the domain is dead-end free, then this strategy will achieve the goal $B_*$ with probability 1.
While this strategy ensures that the robot almost certainly reaches the goal, it incurs a significant computational cost because the robot plans from scratch on each iteration, wasting previous search effort.

\subsection{Reuse-Enforced Replanning}

Although some of the values in the previous plan are no longer viable due to stochasticity,
the plan's overall structure might still be correct. 
Thus, one way to speed up each search is to additionally constrain the next plan to adhere to the same {\em structure} as the previous plan.
To do this, we first identify all action arguments that are {\em constants}, meaning that they are valid quantities in subsequent problems.
These include the names of objects and grasps for objects but not poses, configurations, or trajectories, which are conditioned on the most recent observations of the world. 
We replace each use of a non-constant with a unique free variable symbol (denoted by the prefix \pddl{@}).
For the previous example, this produces the following plan structure, which is used to constrain the second replanning effort: {\small $[\pddl{move}(\pddl{arm}, \pddl{@aq1}, \pddl{@at1}, \pddl{@aq2}),$ $\pddl{pick}(\pddl{D}, \pddl{@p1}, g_\pddl{D}, \pddl{@bq1}, \pddl{@aq2})]$}.
Thus, the planner does not need to search over sequences of actions schemata, objects to manipulate, or grasps because these decisions are fixed.



Algorithm~\ref{alg:policy} gives the pseudocode for our online replanning policy.
The inputs to \method{Policy} are the prior belief $b$, goal set of beliefs $B_*$, and maximum cost $c_*$.
\method{Policy} maintains a set of previously proven static facts $f_{prev}$ as well as the tail of the previous plan $\vec{a}_{prev}$.
On each iteration, first, the procedure \method{Determinize} models the belief SSPP as a deterministic planning problem with actions $A$, initial state $s$, and goal set of states $S_*$.
If the prior plan $\vec{a}_{prev}$ exists, \method{Policy} applies the plan constraints using the \method{ConstrainPlan} procedure described in algorithm~\ref{alg:constrain}. 
If the PDDLStream planner \method{Plan} is unable to solve the constrained problem 
within a user-provided timeout, the constraints are removed, and planning is {\em  reattempted}.
If successful, \method{Plan} returns not only a plan $\vec{a}$ but also the certified facts $f$ within the {\em preimage} of $\vec{a}$ that prove that $\vec{a}$ is a solution.
Then, \method{Policy} executes $a_1$, the first action of $\vec{a}$, receives an observation $o$, and updates its current belief $b$.
Finally, it extracts the subset of constant facts in $f$, static facts that only involve constants, and sets $\vec{a}_{prev}$ to be remainder of $\vec{a}$ that was not executed.


\begin{algorithm}[bt]
    \caption{Online Replanning Policy}
    \label{alg:policy} 
    \begin{algorithmic}[1] 
        \begin{footnotesize}
        \Procedure{Policy}{$b, B_*, c_*$}
            \State $f_{prev}, \vec{a}_{prev} \gets \emptyset, \textbf{None}$
            \While{\textbf{True}}
        		\State $A, s, S_* \gets \Call{Determinize}{b, B_*}$ 
                \State $\vec{a} \gets \textbf{None}$
                \If{$\vec{a}_{prev} \neq \textbf{None}$} \Comment{Reuse plan constraints}
                    \State $A', S_*' \gets \Call{ConstrainPlan}{\vec{a}_{prev}, S_*}$
                    \State $ f, \vec{a} \gets \Call{Plan}{A', s \cup f_{prev}, S_*', c_*}$ 
                \EndIf
        		\If{$\vec{a} = \textbf{None}$} \Comment{No plan constraints}
                    \State $ f, \vec{a} \gets \Call{Plan}{A, s, S_*, c_*}$
        		\EndIf
        		\If{$\vec{a} \neq \textbf{None}$} \Comment{No plan with cost below $c_*$}
        	        \State \Return \textbf{False}
        		\EndIf
        		\If{$\vec{a} = [\;]$} \Comment{Reached goal belief}
        	        \State \Return \textbf{True}
        		\EndIf
    		    \State $o \gets \Call{ExecuteAction}{a_1}$ \Comment{Receive observation $o$}
    		    \State $b \gets \Call{UpdateBelief}{a_1, o}$
    		    \State $f_{prev}, \vec{a}_{prev} \gets \Call{ConstantFacts}{f}, \vec{a}_{2:|\vec{a}|}$
            \EndWhile 
        \EndProcedure
        \end{footnotesize}
    \end{algorithmic}
\end{algorithm}



Algorithm~\ref{alg:constrain} gives the pseudocode for the constraint transformation.
It creates a new set of action schemata $A'$, each of which have modified preconditions and effects, using the previous plan $\vec{a}$.
The fact $(\pddl{Applied}\;i)$ is a total-ordering constraint that enforces that action $a_{i-1}$ be applied before action $a_i$.
For each argument $v$ of action $a_i$, if $v$ is a constant, the new action is forced to use the same value.
The fact $(\pddl{Bound}\;v)$ is true if symbol $v$ has already been assigned to some value in the action sequence. 
If $(\pddl{Bound}\;v)$ is true, the fact $(\pddl{Assigned}\;v\;\pddl{?p})$ is true if free variable $v$ has been assigned to new value \pddl{?p}.
Each free variable $v$ must either be unbound or assigned to action argument \pddl{?p}.


\begin{algorithm}[bt]
    \caption{Plan Constraint Compilation}
    \label{alg:constrain}
    \begin{algorithmic}[1] 
    \begin{footnotesize}
        \Procedure{ConstrainPlan}{$\vec{a}, S_*$}
            \State $A \gets \emptyset$
            \For{$a_i \in \vec{a} = [a_1, a_2, ...]$}
                \If{$2 \leq i$} \Comment{Total ordering constraint}
                    \State $a_i.\text{pre} \gets (\pddlkw{and}\;\;a_i.\text{pre}\;\;(\pddl{Applied}\;\;i-1))$
                \EndIf
                \For{$v, \pddl{?p} \in \textbf{zip}(a_i.\text{args}, a_i.\text{param})$}
                    \If{\Call{IsConstant}{$v$}}  \Comment{Enforce fixed value $\pddl{?p} = v$}
                        \State $a_i.\text{pre} \gets (\pddlkw{and}\;\;a_i.\text{pre}\;\;(\pddlkw{=}\;\; v\;\;\pddl{?p}))$
                    \Else \Comment{Otherwise, $v$ is a free variable}
                        \State $f_p \gets (\pddlkw{imply}\;\;(\pddl{Bound}\;\;v)\;\;(\pddl{Assigned}\;\;v\;\;\pddl{?p}))$
                        \State $a_i.\text{pre} \gets (\pddlkw{and}\;\;a_i.\text{pre}\;\;f_p)$
                        \State $f_e \gets (\pddlkw{and}\;\;(\pddl{Bound}\;\;v)\;\;(\pddl{Assigned}\;\;v\;\;\pddl{?p}))$
                        \State $a_i.\text{eff} \gets (\pddlkw{and}\;\;a_i.\text{eff}\;\;(\pddl{Applied}\;\; i)\;\;f_e)$
                    \EndIf
                \EndFor
                \State $A \gets A \cup \{a_i\}$
            \EndFor
            \State $S_* \gets (\pddlkw{and}\;\;S_*\;\;(\pddl{Applied}\;\;|\vec{a}|))$
            \State \Return $A, S_*$
        \EndProcedure
    \end{footnotesize}
    \end{algorithmic}
\end{algorithm}

\section{Deferred Stream Evaluation}

We use the {\em Focused} algorithm~\cite{garrett2020PDDLStream} to solve each determinized PDDLStream problem.
The {\em Focused} algorithm {\em optimistically} plans using hypothetical stream output values before actually calling any stream procedures.
As a result, it not only generates candidate action plans but also {\em stream plans}, which consist of a sequence of scheduled stream queries that could bind the hypothetical values (\textbf{bolded}) in the action plan.
For example, consider the following possible stream plan for the action plan in section~\ref{sec:replanning}. 

\vspace{-1em}
\begin{small}
\begin{align}\label{eqn:stream}
[&\pddl{grasps}(\pddl{D}) {\to} \pmb{g_\pddl{D}},  \pddl{inv-reach}(\pddl{D}, p^0_\pddl{D}, \pmb{g_\pddl{D}}) {\to} \pmb{q^1_\pddl{b}}, \notag \\
&\pddl{inv-kin}(\pddl{D}, p^0_\pddl{D}, \pmb{g_\pddl{D}}, \pmb{q^1_\pddl{b}}) {\to} \pmb{q^1_\pddl{a}}, \pddl{motion}(\pddl{base}, q^0_\pddl{b}, \pmb{q^1_\pddl{b}}) {\to} \pmb{t^1_\pddl{b}}, \notag \\
&\pddl{motion}(\pddl{arm}, q^0_\pddl{a}, \pmb{q^1_\pddl{a}}) {\to} \pmb{t^1_\pddl{a}}]. 
\end{align} 
\end{small}
\vspace{-1em}

The {\em Focused} algorithm will not terminate until it has successfully bound all hypothetical values. 
As a result, it will recompute the \pddl{motion} stream for every \pddl{move} action on its plan per replanning invocation, spending a significant amount of computation constructing trajectories that will never be executed.
To avoid this, we {\em defer} the evaluation of expensive streams if they are not required before we next replan. 
For the example in equation~\ref{eqn:stream}, the \pddl{inv-kin} and \pddl{motion}(\pddl{arm},...) streams could both be deferred because they are first used within the $\pddl{move}(\pddl{arm}, ...)$ action, rather than the $\pddl{move}(\pddl{base}, ...)$ action.
Still, the stream plan in equation~\ref{eqn:stream} prematurely evaluates \pddl{inv-kin}.
Thus, we {\em reschedule} the stream plan by identifying streams that must be queried to perform the first action or that should never be deferred.
We then {\em recursively} propagate this criterion for streams that are required by the aforementioned streams.
Finally, we query all of the identified streams, which are \pddl{grasps}, \pddl{inv-reach}, and $\pddl{motion}(\pddl{base}, ...)$ and defer the rest: \pddl{inv-kin}, $\pddl{motion}(\pddl{arm},...)$.

It is not always advantageous to defer streams. 
For instance, the initial pose $p^0_\pddl{D}$, sampled grasp $g_\pddl{D}$, and sampled base configuration $q^1_\pddl{b}$ might not admit a kinematic solution (\pddl{Kin}), which is required to perform \pddl{pick}.
Rather than move to $q^1_\pddl{b}$ before discovering this, it is less costly to infer this at the start and sample new values for $g_\pddl{D}$ or $q^1_\pddl{b}$. 
Thus, we only defer the evaluation of streams that are both {\em likely} to succeed and computationally {\em expensive}.
In our domain, we only defer the \pddl{motion} stream as it almost always succeeds as long as the the initial and final configurations are not in collision.




\section{Experiments}

We experimented on 25 randomly generated problems within 4 simulated domains. 
We used PyBullet~\cite{coumans2019} for ray-casting and collision checking and
IKFast~\cite{diankov2010automated} for inverse kinematics.
See {\small \url{https://github.com/caelan/SS-Replan}} for our open-source Python implementation.
We experimented with 4 versions of our system: one per combination of using plan {\em constraints} and/or {\em deferred} streams.
Each policy was limited to 10 minutes of total planning time.

For {the \em Swap} task, the block starts in one drawer, but the goal is to believe it is in the other drawer.
The robot's pose prior is uniform over both drawers.
Successful policies typically inspect the goal drawer, fail to observe it, and then look for it in the other drawer.
This requires placing the block in an intermediate location to close one drawer and open the other.
See the appendix for descriptions of the other tasks. 
Table~\ref{table:results} shows the results of the experiments.
Applying {\em both} plan constraints and deferring streams results in a large improvement in the success rate and generally a reduction in total planning time while executing the policy.

\begin{table}[ht]
\centering
\begin{small}
\begin{tabular}{|l|g|c|g|c|g|c|g|c|}
\hline
\multicolumn{1}{|r|}{\textbf{Alg:}} & \multicolumn{2}{c|}{Neither}&\multicolumn{2}{c|}{Constraints}&\multicolumn{2}{c|}{Deferred}&\multicolumn{2}{c|}{Both}\\
\hline
\textbf{Task:} &
\% & t &
\% & t &
\% & t &
\% & t  \\
\hline
{\em Inspect} & 96 & 41 & \textbf{100} & 45 & \textbf{100} & 32 & \textbf{100} & \textbf{26}
\\ \hline
{\em Stow} & 80 & 124 & \textbf{88} & 226 &\textbf{88} & \textbf{92} & \textbf{88} & 108
\\ \hline
{\em Swap} & 52 & 190 & 40 & 484 & 60 & \textbf{105} & \textbf{80} & 232
\\ \hline
{\em Cook} & 20 & 375 & 40 & 406 & 56 & 249 & \textbf{100} & \textbf{203}
\\ \hline
\end{tabular}
\end{small}
\caption{The success rate (\%) and mean total planning time for {\em successful} trials in seconds (t) over 25 generated problems per task.}
\label{table:results}
\end{table}


We applied our system to real-world kitchen manipulation tasks performed by a Franka Emika Panda robot arm. 
The full perception system is described in prior work~\cite{paxton2019representing}.
See 
{\small \url{https://youtu.be/IOtrO29DFUg}} 
for a video of the robot solving the
the {\em {Inspect Drawer}}, {\em Swap Drawers}, and {\em Cook Spam}
tasks, which are similar to those in table~\ref{table:results}. 

\section{Conclusions}

We presented a replanning system for acting in partially-observable domains.
By planning directly on beliefs, the planner can approximately compute the likelihood of detection given each movable object pose belief.
Through plan structure constraints, we ensure our replanning policy makes progress towards the goal.
And by deferring expensive stream evaluations, we enable replanning to be performed efficiently.



\input{appendix}

\bibliographystyle{IEEEtran}
\bibliography{IEEEabrv,references} 



\end{document}

%% file: appendix.tex
\section*{APPENDIX} \label{sec:appendix}


For each simulated experiment in table~\ref{table:results}, the goal condition, the prior belief, the latent initial state, and a successful execution trace are listed as follows.

\subsubsection{\it Inspect}
The goal is for the green block to be in the bottom drawer and for the bottom drawer to be closed.
The prior for the green block is uniform over both drawers.
The green block is initially in the bottom drawer.
Successful policies open the bottom drawer, detect the green block, and then close the bottom drawer.
The robot intentionally opens the bottom drawer, undoing one of its goals, in order to attempt to localize the green block.
Afterwards, it must reachieve this goal by closing the bottom drawer.

\subsubsection{\it Stow}
The goal is for the green block to be in the top drawer and for the top drawer to be closed. 
The prior for the green block is uniform over the counter, and the prior for the sugar box is uniform over the top drawer.
The green block is initially on the counter, and the sugar box is initially on the top drawer.
Successful policies remove the sugar box from the top drawer (in order to close the top drawer), stow the green block in the top drawer, and close the top drawer.
The robot automatically infers that it must move the sugar box, but not the green block, before closing the top drawer as otherwise the tall sugar box would collide with the cabinet.

\subsubsection{\it Swap}
The goal and prior are the same as in {\em Inspect}.
However, the green block is instead initially in the top drawer.
Successful policies open the bottom drawer, fail to detect the green block, and close the bottom drawer in order to open the top drawer.
Then, they detect the green block, pick up the green block, temporarily place the green block on the counter, close the top drawer, open the bottom drawer, stow the green block in the bottom drawer, and finally close the bottom drawer.
The robot must update its belief upon failing to detect the green block and plan to investigate the other drawer.

\subsubsection{\it Cook}
The goal is for the green block to be cooked.
The prior for the green block is uniform over the counter.
A cracker box and sugar box are initially on the counter, one of which always occludes the green block at its initial pose.
Successful policies move the cracker box and/or the sugar box out of the way until the green block is detected.
Then, they place the green block on the stove, press the stove's button to turn it on (which cooks the green block), and press the stove's button to turn it off.
Depending on the initial pose of the green block and the robot's first manipulation action, the robot might need to inspect behind one or both of the occluding objects in order to localize the spam.